\documentclass[10pt,twocolumn,letterpaper]{article}

\usepackage{iccv}              

\usepackage{makecell}
\usepackage{enumitem} 
\usepackage{tabularx}
\usepackage{booktabs}
\usepackage{multirow} 
\usepackage{makecell} 
\usepackage{graphicx}
\usepackage{float}
\usepackage{utfsym}
\usepackage{pifont}
\usepackage{colortbl}
\usepackage{setspace}
\usepackage{amsmath}
\usepackage{amssymb}
\usepackage{xcolor}
\usepackage{makecell} 
\usepackage{threeparttable}

\definecolor{iccvblue}{rgb}{0.21,0.49,0.74}
\usepackage[pagebackref,breaklinks,colorlinks,allcolors=iccvblue]{hyperref}
\def\paperID{11203} 
\def\confName{ICCV}
\def\confYear{2025}

\title{Rethinking Key-frame-based Micro-expression Recognition: A Robust and Accurate Framework Against Key-frame Errors}

\author{Zheyuan Zhang$^{1,2}$, Weihao Tang$^{3}$, Hong Chen$^{1,2}$\thanks{Corresponding authors.} \\
$^1$Beijing University of Posts and Telecommunications $^3$University of Auckland\\
$^2$Key Laboratory of Interactive Technology and Experience System, Ministry of Culture and Tourism\\
\texttt{\{zzy1998, chenhong76\}@bupt.edu.cn \, wtan402@aucklanduni.ac.nz} 
}

\begin{document}
\maketitle
\begin{abstract}
Micro-expression recognition (MER) is a highly challenging task in affective computing. With the reduced-sized micro-expression (ME) input that contains key information based on key-frame indexes, key-frame-based methods have significantly improved the performance of MER. However, most of these methods focus on improving the performance with relatively accurate key-frame indexes, while ignoring the difficulty of obtaining accurate key-frame indexes and the objective existence of key-frame index errors, which impedes them from moving towards practical applications. In this paper, we propose CausalNet, a novel framework to achieve robust MER facing key-frame index errors while maintaining accurate recognition. To enhance robustness, CausalNet takes the representation of the entire ME sequence as the input. To address the information redundancy brought by the complete ME range input and maintain accurate recognition, first, the Causal Motion Position Learning Module (CMPLM) is proposed to help the model locate the muscle movement areas related to Action Units (AUs), thereby reducing the attention to other redundant areas. Second, the Causal Attention Block (CAB) is proposed to deeply learn the causal relationships between the muscle contraction and relaxation movements in MEs. Empirical experiments have demonstrated that on popular ME benchmarks, the CausalNet has achieved robust MER under different levels of key-frame index noise. Meanwhile, it has surpassed state‑of‑the‑art (SOTA) methods on several standard MER benchmarks when using the provided annotated key‑frames. Code is available at \url{https://github.com/tony19980810/CausalNet}.
\end{abstract}

\section{Introduction}
\label{sec:intro}

\begin{figure}[!t]
\centering
\includegraphics [width=1\linewidth]{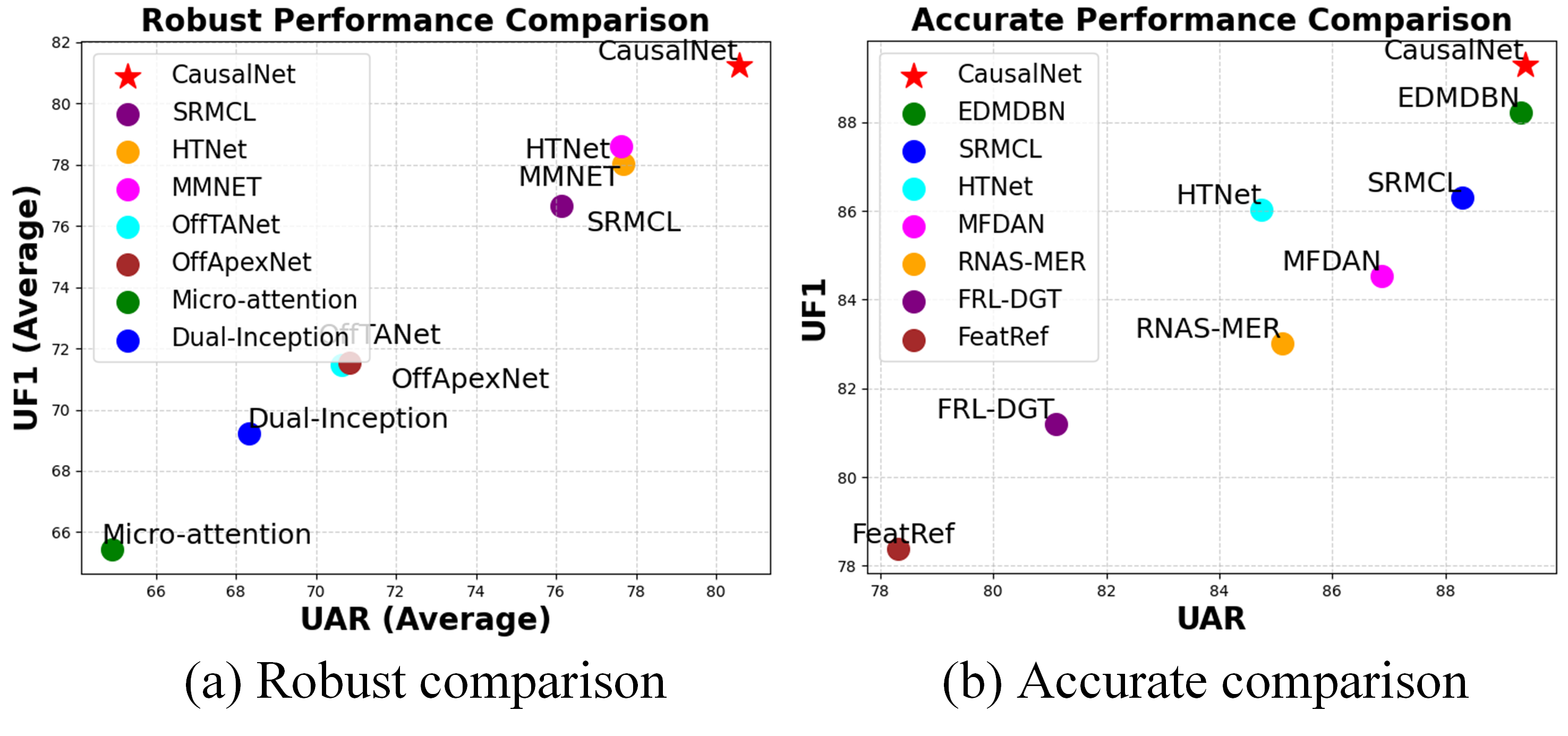}
\caption{Compared with existing advanced methods, CausalNet ranks first in both robustness and accuracy assessment on the composite dataset including SAMM \cite{davison2016samm}, CASME II \cite{yan2014casme} and SMIC \cite{li2013spontaneous}. The robustness assessment is based on the mean values of UAR and UF1 under three different levels of key-frame index errors (see \cref{s4.2}), while the accuracy assessment is through the standard evaluation with the annotated key-frames (see \cref{s4.3}).}
\label{fa1}
\end{figure}

Expression is the manifestation of emotion directed toward given stimuli \cite{zhou2024seeing,liu2025human,zhang2025masked,10.1007/978-981-96-5809-1_20,10822714}. In the real world, expressions can be roughly divided into two types: macro-expressions and micro-expressions (MEs). The former exhibits strong facial muscle movements for a long duration and is relatively easy to recognize. The latter is an involuntary, fleeting, and unconscious facial expression that occurs when a person fails to control their facial expressions. Therefore, the emotion conveyed by such expressions is relatively more genuine \cite{11060844,s25092866}. Nevertheless, when MEs occur, the muscle movements are weak, and the duration is short (less than 0.5 seconds \cite{ekman1971constants}), which poses a huge challenge for both humans and machines to recognize. Therefore, accurately spotting and recognizing MEs is of great significance in affective computing \cite{li2022deep, 10889587,zhou2024ulme, liu2025lightweight}.

Onset, apex, and offset frames correspond to the start, peak, and end of the muscle movements in MEs respectively. Based on the locations of these key-frames, key-frame-based methods with the core idea of “less is more" \cite{LIONG201882} filter out redundant frames without obvious expression information and take partial ME frames as input (e.g., only the apex frame \cite{wang2020micro, 8756544} and the onset-apex dynamic representation \cite{cai2024mfdan, zhang2021off, fan2023selfme, nguyen2023micron, bao2024boosting, wang2020micro, ijcai2022p150, verma2023rnas, zhai2023feature, ZHOU2022108275,wang2025cross,10610258}). Owing to the extraction of crucial ME information, these methods have propelled the accuracy of MER to an entirely new level. However, despite the significant improvement in the accuracy of MER, the recognition performance in the practical applications of MEs has not been enhanced to the same extent. This has led us to rethink key-frame-based MER: In the research on improving MER performance, is the default use of relatively accurate key-frames annotated by experts for MER in line with the requirements of practical applications? Has the great difficulty in obtaining accurate key-frame locations and the impact of key-frame errors on MER been overlooked? Regardless of the answers to the above questions, an objective fact is that even among different experienced experts, because of individual biases, their key-frame annotation results for the same ME sample will still show certain differences. This implies that key-frame errors are always present, whether through manual annotation or by using automatic spotting algorithms. To quantify and evaluate the impact of errors on existing methods, we conduct experiments on the composite dataset of three ME benchmarks \cite{davison2016samm,li2013spontaneous,yan2014casme} through two evaluation approaches: manually introducing different levels of errors based on the annotated key-frames provided by the datasets and using the spotted indexes by the automatic spotting algorithm. Experimental results show that the recognition performance of existing key-frame-based methods is severely affected (see \cref{s4.2} for details). The low robustness of these methods hinders their practical applications. 

In this paper, we focus on key-frame-based MER and propose CausalNet, a framework that is robust to key-frame index errors and can maintain accurate MER at the same time as shown in \cref{fa1}. In terms of robustness, CausalNet uses the onset-offset full ME sequence representation (onset-apex optical flow (OF) and apex-offset OF) as the input to achieve robust MER. Compared with partial ME input like onset-apex frames or only apex frames, a complete representation of MEs has better robustness in capturing the key information of facial muscle movement. In terms of accurate MER, we design two parts of networks: First, to enable the model to focus on the key areas and reduce attention to redundant information, we develop a causal motion position learning module. It provides the model with the position information of muscle movement by learning the causal relationship between the directions of muscle contraction and relaxation in the onset-apex and apex-offset phases. Second, to deeply explore the causal relationships between onset-apex and apex-offset MEs and address the weakening of the model's perception ability of local features as feature fusion deepens \cite{peng2021conformer}, we propose the causal attention block. Its unique design enables the model to perceive the overall temporal trend of MEs while maintaining sensitivity to local key ME information.

In brief, the core contributions of our work are:
\begin{itemize}[label=\textbullet] 
    \item We propose a novel framework using onset-apex OF and apex-offset OF as inputs to achieve robust and accurate MER facing key-frame index errors.
	\item We conduct a Causal Motion Position Learning Module (CMPLM) to enable the model to focus on the key muscle movement areas in different ME stages while ignoring other redundant information.
    \item We present a Causal Attention Block (CAB) to learn causal relationships of muscle movements between onset-apex and apex-offset ME phases. The unique design helps the model maintain sensitivity to local key information.
    \item We demonstrate the robustness of the proposed method against key-frame errors of different levels. Meanwhile, on standard ME benchmarks, the proposed method shows competitive performance.
\end{itemize}

\section{Related Work}
\begin{figure*}[t]
\centering
\includegraphics [width=1\linewidth]{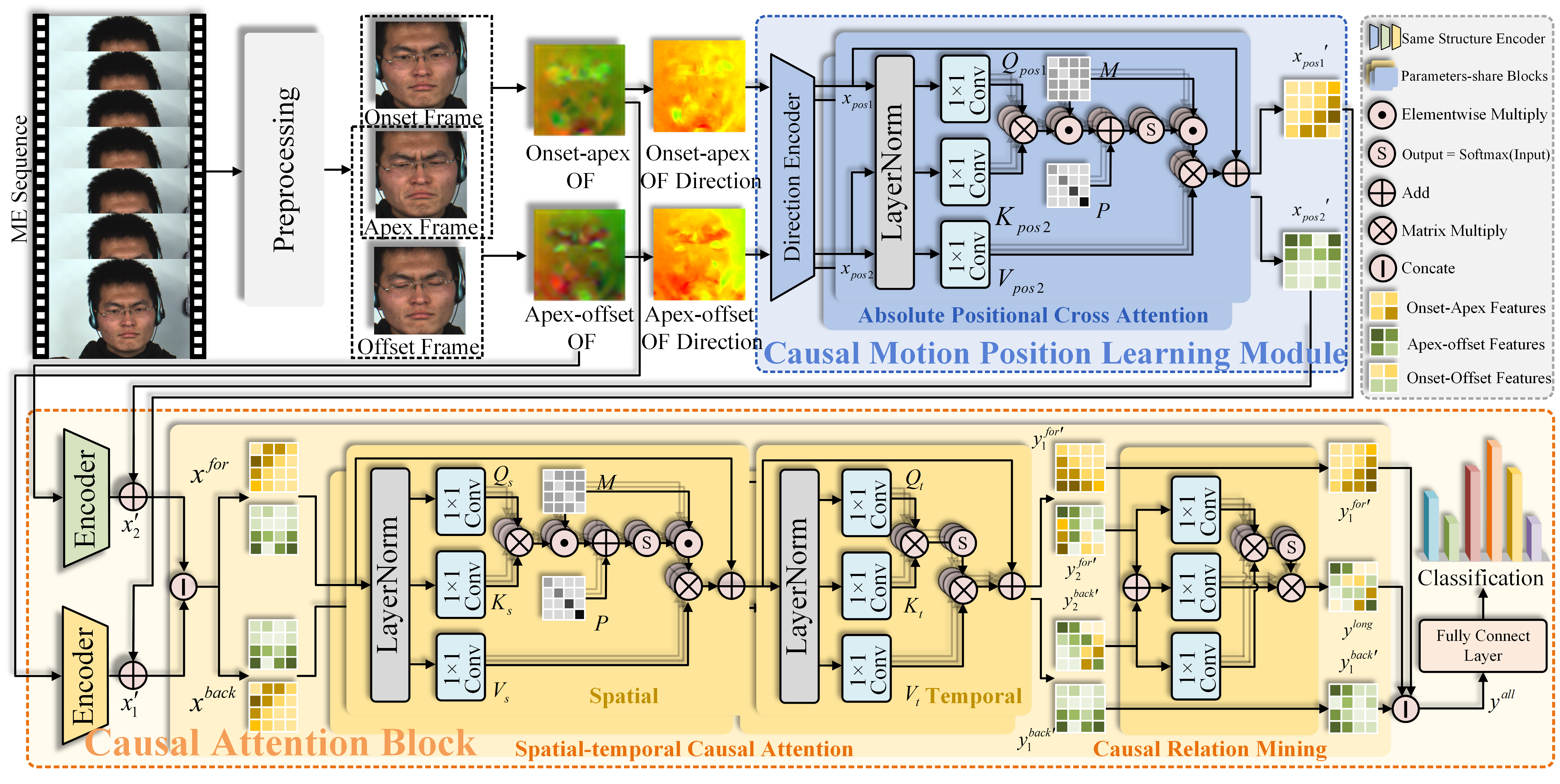}
\caption{The overview of the proposed CausalNet with two main parts: the CMPLM (blue part) and the CAB (yellow part). CMPLM takes the directional components of the two OFs as input. Through absolute positional cross attention, it obtains the AU-related position information of the OFs and feeds them back to the CAB in the form of position embedding. The CAB has two components: (1) Spatial-temporal causal attention conducts spatio-temporal information interaction on the bi-directionally concatenated onset-apex (yellow features) and apex-offset features (green features). Due to the causal design in the temporal dimension, short-range (pure green and pure yellow features) and long-range feature representations (yellow-green mixed features) are obtained. (2) Causal relation mining enables information interaction among long-range features. Finally, the fully-connected layer classifies the concatenated features.}

\label{f4}
\end{figure*}

\noindent\textbf{Key-frame-based MER}. According to the input, these methods can be roughly divided into three categories: (1) Apex-frame-based MER: Wang \etal \cite{wang2020micro} propose micro-attention which guides the network to focus on crucial expression areas of apex frame for MER. (2) Onset-apex-frame-based MER: Wang \etal \cite{wang2024htnet} design a hierarchical transformer network using multi-scale onset-apex OF features, and Nguyen \etal \cite{nguyen2023micron} develop an unsupervised MER training mechanism. Yet, due to the restricted ME sequence representation range, these two types of methods are not robust against key-frame errors. (3) Onset-offset-frame-based MER: Zhu \etal \cite{zhu2023learning} split ME video clips, randomly select frames from segments, and combine them with onset and offset frames to represent MEs. The third kind of approach provides a more comprehensive ME representation and better resistance to key-frame errors compared to the previous two. CausalNet belongs to this category but differs from existing methods by addressing information redundancy and implementing causal relationship learning of MEs, enabling more accurate MER.

\noindent\textbf{Video-based MER} takes video clips of ME sequence as input. Zhang \etal \cite{9915457} propose a spatio-temporal transformer to learn ME relations in different ranges. Liu \etal \cite{10687556} propose a self-supervised MER method via a temporal Gaussian masked autoencoder, solving the problem of scarce labeled data. Similar to the onset-offset-based approach, video-based methods are relatively robust to key-frame index errors because of the complete ME input. However, they can still be influenced because most of these methods use key-frame indexes to crop or interpolate inputs (e.g., Zhang \etal \cite{9915457} interpolate with onset, apex, and offset frame indexes, and Liu \etal \cite{10687556} select 16 frames as input).
Full-input methods face MER performance limitations and high computational costs due to redundant information and large data volume. Compared with the video-based methods, CausalNet is not less robust than video-based methods to key-frame index errors. In addition, it effectively tackles information redundancy and shows stronger MER performance.

Our study represents an initial endeavor to explore the enhancement of robustness in key-frame-based MER against key-frame errors. The proposed method aims to maintain robust and accurate MER, both in the absence and presence of key-frame errors.

\begin{figure*}[t]
\centering
\includegraphics [width=1\linewidth]{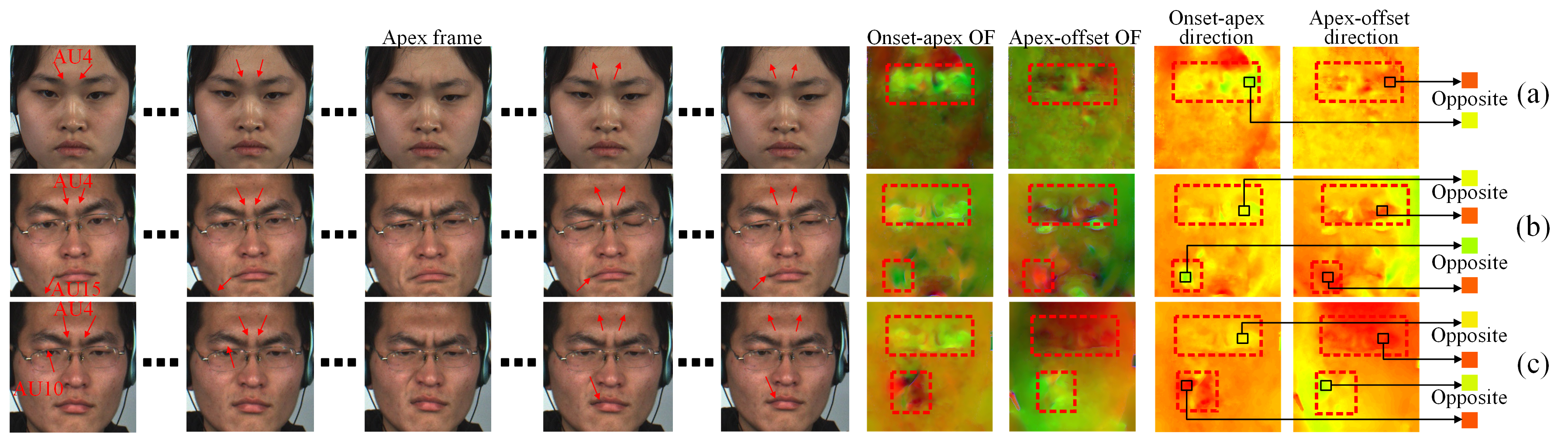}
\caption{ME sequence, corresponding OFs and OF direction maps. We use different hues to represent different angles of motion directions of OFs in OF direction maps. In (a), the muscle movement directions of AU4 in onset-apex and apex-offset phases are basically opposite (In our HSV color space, red roughly represents the upward direction and yellow-green represents the downward one). The same is true for AU4 and AU15 in (b), and for AU4 and AU10 in (c). See Supplementary Materials for more visualizations.}
\label{f0}
\end{figure*}

\section{The Proposed CausalNet}

Enhancing the robustness against key-frame index errors is a significant focus of CausalNet. As shown in \cref{f4}, CausalNet achieves robust MER by using onset-apex and apex-offset OFs to represent the whole ME sequence. To mitigate the focus of the model on redundant information and maintain accurate MER, CMPLM is proposed to learn AU-related position information through the OF direction maps, thus reducing attention to non-expression areas. In the CAB, the spatial-temporal causal attention is responsible for the information interaction. Meanwhile, based on the causality in the temporal dimension, it generates short-range and long-range feature representations, so as to perceive the overall temporal trend of MEs while maintaining sensitivity to local key ME information. Causal relation mining enhances causal relationship learning through in-depth information interaction and mining of long-range features.

\subsection{Causal Motion Position Learning Module}

\noindent\textbf{Causal Relations in MEs}. As shown in \cref{f0}, we separately calculate the motion direction from the OFs, and visualize the OF motion directions using different hues. The visualized colors of the muscle movements in the AU regions (shown within the red solid-line boxes) show that the motion directions of muscles between the onset-apex and apex-offset phases are almost opposite. This indicates that the muscle movements in MEs follow specific patterns. During the onset-apex phase, the muscles contract, while during the apex-offset phase, the muscles relax. The directions of muscle movements in these two phases are almost opposite, forming a causal relationship.

\begin{figure}[t]
\centering
\includegraphics [width=1\linewidth]{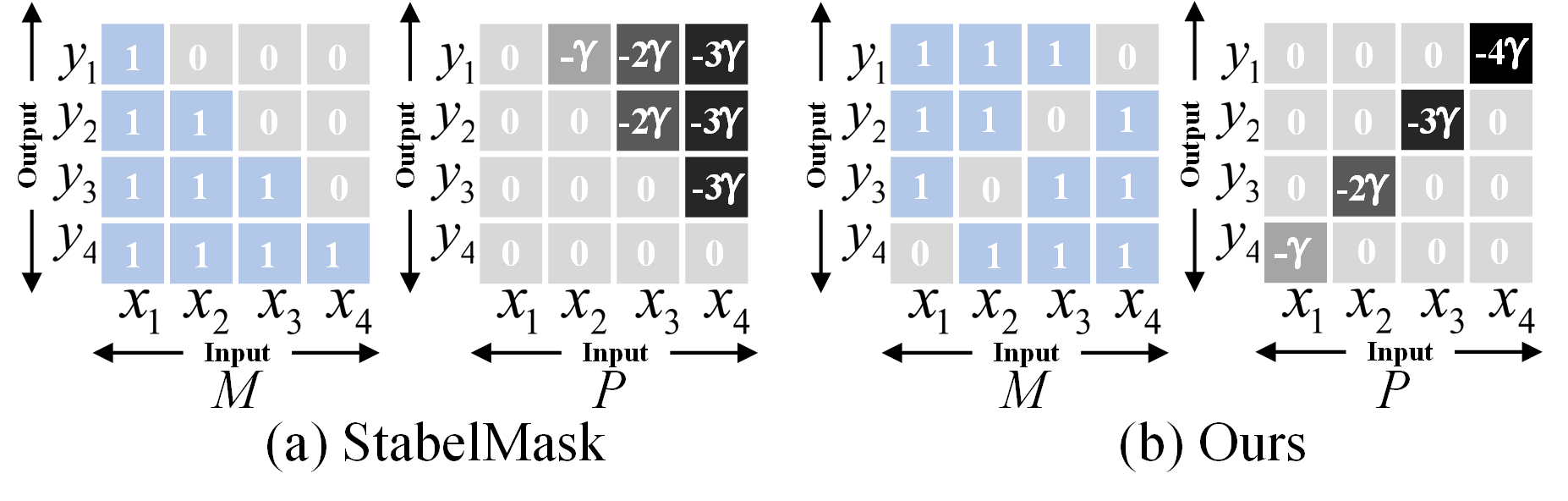}
\caption{Visualizations of $M$ and $P$ in StableMask \cite{yin2024stablemask} (a) and absolute position cross attention (b).}
\label{MP}
\end{figure}

Building upon this discovery, we devise the causal motion position learning module, which takes the motion directions of the onset-apex and apex-offset OFs as inputs. The aim is to locate the AU-related muscle movement regions by perceiving the changes in the motion directions of AU-related muscles during the contraction and relaxation phases. Subsequently, through position embedding, this valuable information is channeled to the main branch of the model. This empowers the model to concentrate its attention on the AU-related muscle movements, effectively filtering out other redundant information. The module consists of two distinct components: The direction encoder is responsible for feature extraction of the OF direction map, and the absolute positional cross attention enables information interaction between the two OF direction features, so as to keenly perceive the location details of the AU-related muscle movement regions.

\noindent\textbf{Introducing Absolute Position Information}. Since the input face images are all cropped and aligned, the approximate regions of AU positions in the images are relatively fixed. In this scenario, the absolute position information of image patches enables the model to precisely identify the specific facial regions corresponding to different patches. The introduction of it will empower the model to more effectively capture the spatial relationships among facial features, thereby enhancing the performance of MER. To achieve this goal, inspired by StableMask \cite{yin2024stablemask} in natural language processing, we introduce pseudo-attention scores which converts the attention matrix into a non-right stochastic matrix (the sum of the elements in each row is no longer equal to 1) to encode the absolute position information.

\noindent\textbf{Absolute Positional Cross Attention} uses a sparse attention mechanism for local information interaction. It adds pseudo-attention scores to the masked positions of the attention score matrix to encode absolute positional information. Specifically, after feature extraction by the direction encoder on two OF direction maps, two feature $x_{pos1}, x_{pos2}\in\mathbb{R}^{m^2\times{D}}$ is obtained. $m$ stands for the resolution of the feature map and is equal to 2, while $D$ denotes the feature dimension. For $x_{pos1}$ (same for $x_{pos2}$), the absolute positional cross attention can be expressed as: 
\begin{equation}
\text{PosAttn}(Q_{pos1}, K_{pos2},V_{pos2}) = \widetilde{A}_{pos}{V_{pos2}},
\end{equation}
\begin{equation}
\widetilde{A}_{pos}=\text{SM}(\frac{Q_{pos1}K_{pos2}^{T}}{\sqrt{d_k}}\odot M + P)\odot M,
\label{e1}
\end{equation}
\begin{equation}
M(i,j)=
\begin{cases}
1, \text{if } (x_i-x_j)^2+(y_i-y_j)^2\leq r\\
0, \text{otherwise}
\end{cases},
\end{equation}
\begin{equation}
P(i,j)=
\begin{cases}
0, \text{if } (x_i-x_j)^2+(y_i-y_j)^2\leq r\\
-j\gamma, \text{otherwise}
\end{cases}.
\end{equation}
where $K_{pos2}$ and $V_{pos2}$ are obtained through a 1×1 convolution of $x_{pos2}$, while $Q_{pos1}$ is obtained via a 1×1 convolution of $x_{pos1}$. $\sqrt{d_k}$ serves as the scaling factor. SM is the softmax function. $\odot$ represents element-wise multiplication, and $M$ is a two-valued mask matrix to mask out certain locations according to the relative position of \textit{query} (at $i$) and \textit{key/value} (at $j$) tokens. $(x_i, y_i)$ and $(x_j , y_j)$ are the spatial locations of $i$ and $j$. $P$ is the two-valued matrix containing pseudo-attention score. Visualizations of $P$ and $M$ are shown in \cref{MP} (b). $\gamma$ is a positive decay rate hyperparameter. $r$ is the neighborhood radius and equal to 1. After the absolute positional cross attention, ${x_{pos1}}', {x_{pos2}}'\in\mathbb{R}^{m^2\times{D}}$ are obtained.

\noindent\textbf{How Pseudo-attention Score Encodes Absolute Position}. Since the original pseudo-attention is used together with the causal mask (first of \cref{MP} (a)), to adapt to the sparse attention, we made changes to the $P$ (second of \cref{MP} (b)). Here, we prove that the modified $P$ can still encode absolute position information using the same example in StableMask \cite{yin2024stablemask}: whether the model can encode positional information for an identical input sequence. let $A_{i}$ denote the real attention scores of the $i-$th row ($A=QK^{T}/\sqrt{d_k}$), $P_{i}$ denote the pseudo-attention scores in the $i -$th row , and $j$ means the $j -$th column, we have:
\begin{equation}
\sum \text{SM}_{A_{i} \cup P_{i}}\left(A_{i}\right)=1-\sum \text{SM}_{A_{i} \cup P_{i}}\left(P_{i}\right),
\end{equation}
where $\text{SM}_{A_{i} \cup P_{i}}(A_{i})$ and $\text{SM}_{A_{i} \cup P_{i}}(P_{i})$ are the real/pseudo attention in each row. For an identical input sequence $X=[x,x,x,x]$ (in our scenario, $X$ is an OF image that contains 4 patches $x$), $\sum_{i \neq j} \exp(A_{i,j})$ remains constant as $i$ increases as each row has three equal real attention score because of identical input (our $M$ masks one query per row ), and $\sum_{i=j} \exp(P_{i,j})$ for pseudo attention in mask position increases as $i$ increases ($\exp(-4\gamma)<...<\exp(-\gamma)$), we have
\begin{equation}
\sum \text{SM}_{A_{i} \cup P_{i}}\left(A_{i}\right)>\sum \text{SM}_{A_{i + 1} \cup P_{i+1}}\left(A_{i+1}\right),
\end{equation}
which means after \cref{e1}, the output attention values become monotonic instead of being all the same: 
\begin{equation}
\widetilde{A}(W_{V}X)^{\top}=\left[\alpha_{1}v,\alpha_{2}v,\alpha_{3}v,\alpha_{4}v\right],
\end{equation}
\begin{equation}
1>\alpha_{1}>\alpha_{2}>\alpha_{3}>\alpha_{4}>0.
\end{equation}
where $W_{V}$ is the weight matrix of $V$, and $v$ is a single vector within $V$. This indicates that $P$ in our method is consistent with the properties of StableMask. Different absolute positions have different sums of attention score, and the sum of attention score changes monotonically as the absolute position increases, thereby encoding the absolute positions of an identical input sequence.

\subsection{Causal Attention Block}
\label{cab}

\begin{figure}[t]
\centering
\includegraphics [width=1\linewidth]{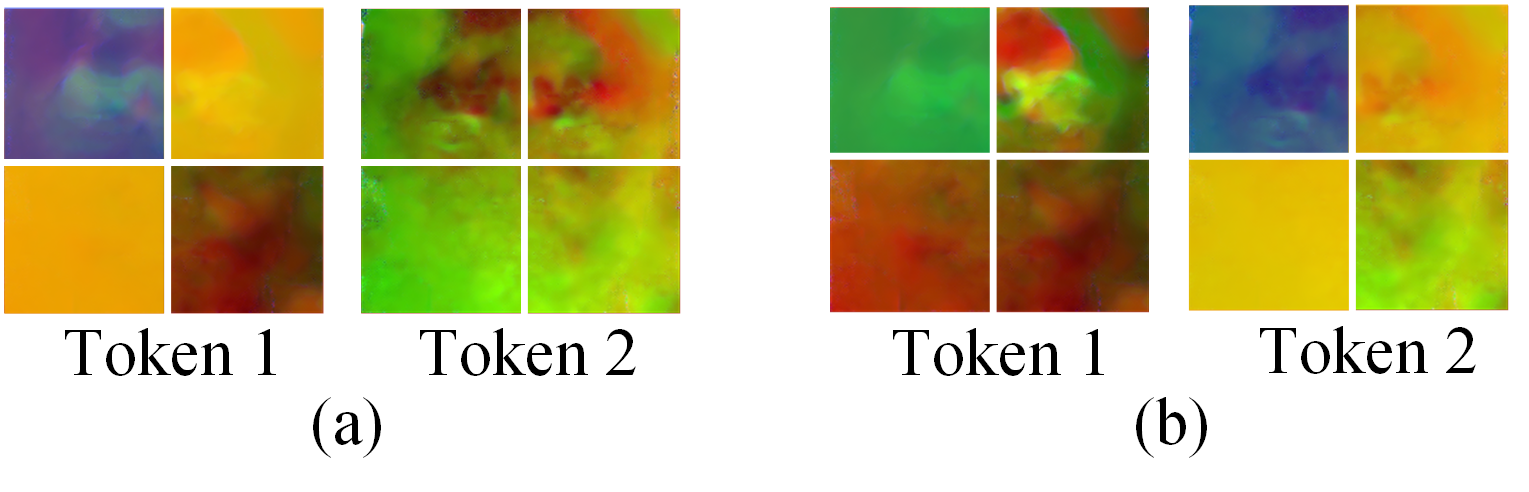}
\caption{Visualization of the proposed spatial-temporal causal attention on forward-concatenated features. (a) for token 1 (onset-apex OF) attention calculation, (b) for token 2 (apex-offset OF) attention calculation. For illustration, we denote the query patch in blue, and show the patches participating in spatial attention in yellow, and those involved in temporal attention in green.}
\label{causal}
\end{figure}

\noindent\textbf{Spatial-temporal Causal Attention}. As shown in \cref {causal}, this module first conducts sparse spatial information interaction, and its mechanism is consistent with that of the absolute positional cross attention. Then, the module performs causality-based information interaction in the temporal dimension to generate short-range and long-range representations. Compared with self-attention, due to causality, it ensures that as the feature fusion deepens, the short-range features only focus on short-range information.

 Specifically, after feature extraction by the encoder from two OF images, a feature $x\in\mathbb{R}^{2\times {m^2\times{D}}}$ composed of two tokens $x_{1,2}'\in\mathbb{R}^{{m^2\times{D}}}$ is obtained. After adding the positional information from CMPLM, the two tokens are concatenated bidirectionally to get $x^{for}$ and $x^{back}$ as the input for spatial-temporal causal attention. The purpose of the bidirectional features are to generate two short-range features that focus on the apex-offset range and the onset-apex range and two long-range features of the onset-offset range. The formula for spatial attention is as follows:

\begin{equation}
\text{SpatialAttn}(Q_s, K_s, V_s) = \widetilde{A}_{s}{V_{s}},
\end{equation}
\begin{equation}
\widetilde{A}_{s}=\text{SM}(\frac{Q_{s}K_{s}^{T}}{\sqrt{d_k}}\odot M + P)\odot M.
\end{equation}
where for the \textit{query} (at $i$) and \textit{key/value} (at $j$) tokens, $t_i=t_j$. $t_i,t_j \in \{1, 2\}$ represents the temporal dimension for the tokens.

\begin{figure*}[t]
\centering
\includegraphics [width=1\linewidth]{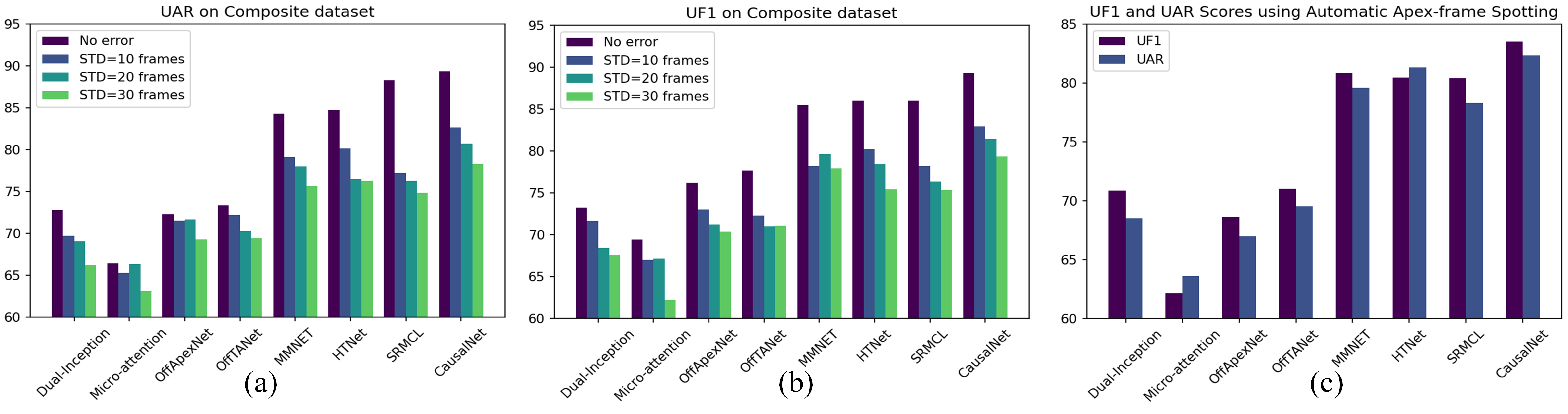}
\caption{Results of robust MER evaluation under key frames with different levels of errors on the composite dataset (a and b) and automatic key-frame spotting on CASME II (c). The compared methods are Dual-Inception \cite{8756579}, Micro-attention$^{*}$ \cite{wang2020micro}, OffApexNet$^{*}$ \cite{GAN2019129}, OffTANet$^{*}$ \cite{zhang2021off}, MMNET$^{*}$ \cite{ijcai2022p150},  HTNet \cite{wang2024htnet}, SRMCL \cite{bao2024boosting} and CausalNet. $*$ represents that for these methods, relevant results of CDE task with annotated key-frames have not been reported in their paper. So we reproduce the results based on the provided code. }
\label{f5.5}
\end{figure*}

For efficient temporal attention, we let tokens at the same spatial positions interact as each token already holds global information after spatial interaction. The interaction adheres to causal attention, meaning previous tokens in the temporal dimension don't incorporate information from subsequent ones. Specifically, we only calculate attention for the second temporal-dimension token, and the formula is as follows:

\begin{equation}
\text{TemporalAttn}(Q_{t}, K_{t},V_{t}) = \text{SM}(\frac{Q_{t}K_{t}^{T}}{\sqrt{d_k}}){V_{t}}.
\end{equation}
where for the \textit{query} (at $i$) and \textit{key/value} (at $j$) tokens,  $(x_i,y_i)=(x_j,y_j)$, $t_j\in \{1, 2\}$  and $t_i\in \{ 2\}$ because the above-mentioned temporal attention is only applied to the second OF feature to ensure causality. In the end, ${{y}^{for}}', {{y}^{back}}'\in\mathbb{R}^{2\times m^2\times D}$ is obtained and divided into ${y^{for}_{1}}'$, ${y^{for}_{2}}'$, ${y^{for}_{2}}'$, ${y^{back}_{2}}' \in\mathbb{R}^{m^2\times D}$. Due to causality, ${y^{for}_{1}}'$ and ${y^{back}_{1}}'$ are the short-range features focus on onset-apex and apex-offset OF respectively, while ${y^{for}_{2}}'$ and ${y^{back}_{2}}'$ are the long-range features focus on onset-offset OF.

\noindent\textbf{Causal Relation Mining} is proposed to deeply explore the relation between the contraction and expansion of ME in the long-range features through information interaction. In terms of formulation, the process is as follows:

\begin{equation}
y^{long}=\text{SM}(\frac{{y^{for}_{2}}'{{y^{back}_{2}}}'^{T}}{\sqrt{d_k}})({y^{for}_{2}}'+{y^{back}_{2}}').
\end{equation}
where $y^{long}\in\mathbb{R}^{{m^2\times{D}}}$. In the end, $y^{long}$, $y^{for}_{1}$ and $y^{back}_{1}$ are concatenated together to form $y^{all}\in\mathbb{R}^{{3\times m^2\times{D}}}$  and sent to a fully-connected layer for classification.

\section{Experimental Results}

\subsection{Datasets, Protocols and Implementation Details}
 \noindent\textbf{CASME II} \cite{yan2014casme}. Captured at a sampling rate of 200 frames per second, it comprises 247 ME samples from 26 individuals of the same ethnicity. 

 \noindent\textbf{SAMM} \cite{davison2016samm}. Recorded at a frame rate of 200 frames per second, the dataset features 159 samples from 32 participants across 13 different ethnicities. 

 \noindent\textbf{SMIC} \cite{li2013spontaneous} contains 164 samples and includes participants from 16 different ethnicities, with recordings from 3 participants. The frame rate is 100 frames per second.

 \noindent\textbf{MMEW} \cite{9382112} includes 300 ME sequences from 36 participants with a frame rate of 90 frames per second.

 \noindent\textbf{Protocols}. To facilitate comparison and assessment, this work adheres to the standards set by MEGC 2019 \cite{see2019megc} and conducts three-class Composite Database Evaluation (CDE) on the first three datasets and three-class and five-class evaluation on MMEW (with extra training data from CDE task) under the Leave-One-Subject-Out (LOSO) evaluation protocol. The validation metrics used are the unweighted F1 score (UF1), unweighted accuracy (UAR) and accuracy (ACC). 

 \noindent\textbf{Implementation Details}. The OF images (with a size of 28×28×3) consist of three channels (horizontal, vertical, and optical strain elements) \cite{wang2024htnet}. The experiments are conducted using PyTorch 2.2.1 and Python 3.8. In the training stage, we set the learning rate to 5e-5 and use a maximum of 800 epochs with the adaptive moment estimation optimizer \cite{kingma2014adam}. $\gamma$ can be referred to in the code link, and we employ HTNet \cite{wang2024htnet} for all encoders. 

\subsection{Robust MER with Inaccurate Key Frames}
\label{s4.2}

\begin{figure*}[t]
\centering
\includegraphics [width=1\linewidth]{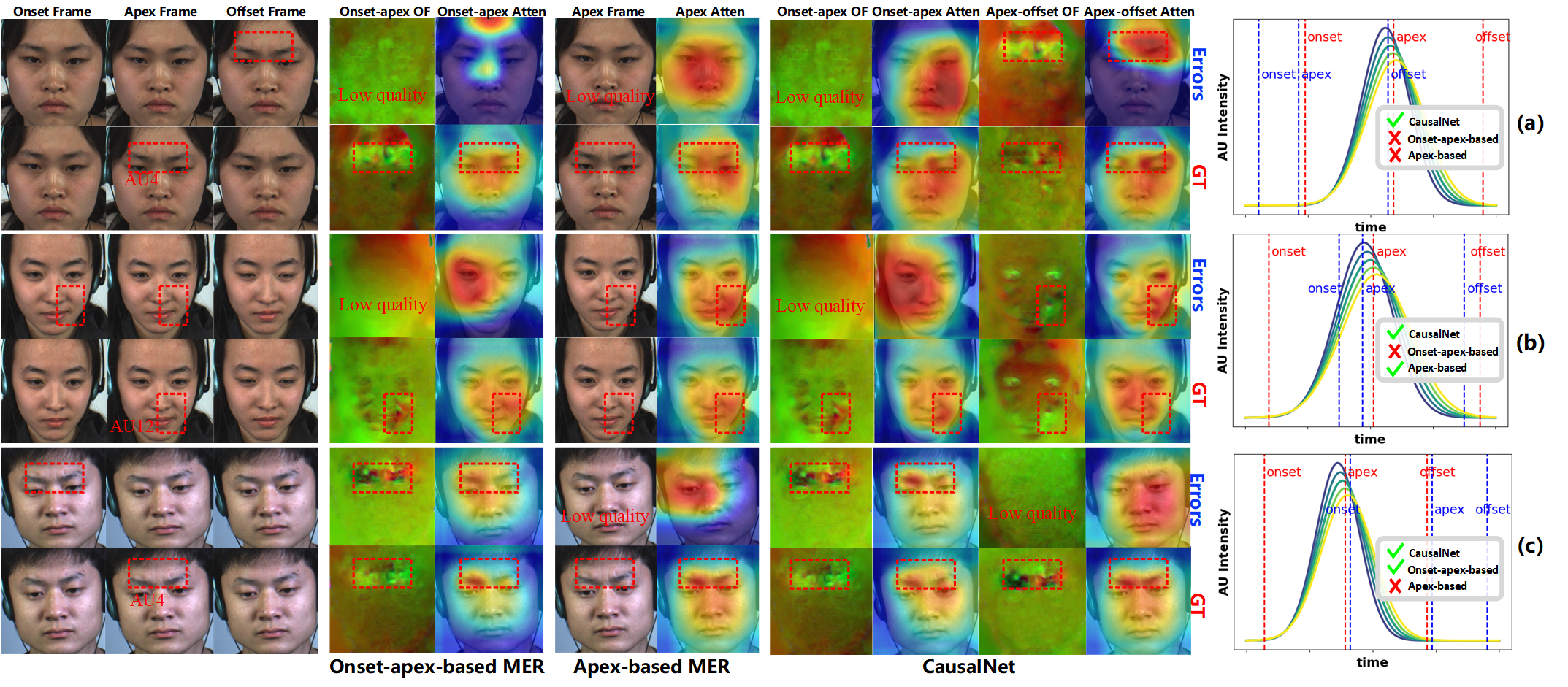}
\caption{Visualization of OFs and corresponding attention maps \cite{selvaraju2017grad} in ground truth (GT) key-frames ({\color{red}{\textbf{red}}}) and key-frames with errors ({\color{blue}{\textbf{blue}}}) on the three kind of key-frame-based method. On the right side is the visualization of the GT and the key-frames with errors on the time axis. For better visualization, we set the backbone to ResNet18 \cite{he2016deep} and the input size of the OFs to 224×224×3. The AU regions related to facial expressions have been marked with red boxes. Situations where there is less facial expression information in the apex frame or OF due to key-frame errors have been labeled as “low quality".}
\label{f5}
\end{figure*}

\noindent\textbf{Error Setting}. We assume key-frame errors follow a normal distribution centered on the datasets' manually annotated key-frame indexes. We set standard deviations (STDs) of 10, 20, and 30 frames and introduce them to apex, onset, and offset frames on the composite datasets to analyze performance changes of existing open-source state-of-the-art (SOTA) key-frame-based methods \cite{8756579,wang2020micro,GAN2019129,zhang2021off,ijcai2022p150,wang2024htnet,bao2024boosting} and CausalNet. Given SMIC's half data acquisition frame rate of CASME II and SAMM, its STD is halved. It is worth noting that MMEW is not involved in this experiment. In addition to the manual error addition, we employ the apex frame spotting algorithm \cite{liu2023long} for apex frame localization to introduce key-frame errors from automatic detection (single database evaluation using CASME II is conducted here). The sequence range for spotting is the annotated onset-offset sequences, where the predicted onset and offset frames are 25 frames apart from the predicted apex frame.

\noindent\textbf{Robustness Evaluation under Key-frames with Errors}. In \cref{f5.5} (a) and (b), for manually-introduced errors on the composite dataset, the results show that CausalNet is in the leading position under the metrics of all three STDs. Specifically, when the STD is 10, 20, and 30 frames respectively, compared with the second-best method, the UF1 of CausalNet has improvements of 2.70\%, 1.85\%, and 1.42\% respectively. In terms of the UAR, there are improvements of 2.47\%, 2.70\%, and 2.01\% respectively. In \cref{f5.5} (c), for automatic apex-frames spotting on CASME II, CausalNet also achieves the best result with improvements of 2.62\% and 1.06\% for UAR and UF1 respectively, which demonstrates better robustness against key-frame errors.

\noindent\textbf{Analysis of How CausalNet Achieves Robust MER}. We have listed three situations where the apex-based and onset-apex-based methods are significantly affected by errors, as shown in \cref{f5}.  In (a), the ME information is completely outside the onset-apex range, resulting in the onset-apex OF having no muscle information of AU 4. This will have a huge impact on the methods based only on the apex frame or the onset-apex OF. However, CausalNet encompasses a more comprehensive ME motion range, making it more probable to incorporate the peak muscle movement process. The muscle movement information lost is compensated in the apex-offset OF. In (b), since the onset frame with errors is too close to the apex frame, the relative motion between the two frames is not obvious. As a result, the quality of the calculated onset-apex OF is poor, which in turn affects the onset-apex-based MER. However, the apex-frame-based method and CausalNet show good robustness in this situation. In (c), compared with the ground truth, the key-frames with errors are shifted backward as a whole. As a result, there is less facial expression information in the apex frame, which has a significant impact on the apex-based MER, while the onset-apex-based method and CausalNet remain robust in this situation. In addition, the situations where CausalNet is more robust to errors than other key-frame-based MER methods are not limited to the above three cases. The more comprehensive ME range input enables CausalNet to have better robustness.

\begin{table*}[t]\centering

  \caption{Method Comparison on the CASME II, SMIC, SAMM and the composite dataset. The results of CausalNet are calculated from 5 independent runs, with the subscript representing the STD, and $\pm$ is omitted.
Other results are reported from the corresponding papers, and we highlight for each of them the {\color{red}{\textbf{best}}} and {\color{blue}{\textbf{second-best methods}}}. $\dagger$ indicates these methods are video-based methods. $*$ indicates that the relevant results are only retained to one decimal place in the corresponding papers. - indicates that the results are not reported in the corresponding papers.}
  \label{t1}
  \setstretch{1.3}
  \resizebox{\linewidth}{!}{
  \begin{tabular}{cccccccccccccc}
    \toprule
    \multirow{2}{*}{\textbf{Method}}&\multirow{2}{*}{\textbf{Pub-Yr}}&\multicolumn{3}{c}{\textbf{CASME II}}&\multicolumn{3}{c}{\textbf{SMIC}}&\multicolumn{3}{c}{\textbf{SAMM}}&\multicolumn{3}{c}{\textbf{Composite}}\\
    &&\textbf{UF1}&\textbf{UAR}&\textbf{ACC}&\textbf{UF1}&\textbf{UAR}&\textbf{ACC}&\textbf{UF1}&\textbf{UAR}&\textbf{ACC}&\textbf{UF1}&\textbf{UAR}&\textbf{ACC}\\
    \toprule
    FeatRef \cite{ZHOU2022108275}&PR 22&89.15&88.73&-&70.11&70.83&-&73.72&71.55&-&78.38&78.32&-\\
 SLSTT-LSTM$^{*\dagger}$ \cite{9915457}&TAC 22&90.1&88.5&-&74.0&72.0&-&71.5&64.3&-&81.6&79.0&-\\
    FRL-DGT$^{*}$ \cite{zhai2023feature}&	CVPR 23&91.9	&90.3&-&	74.3&	74.9&-&	77.2&	75.8&-&	81.2&	81.1&-\\
    RNAS-MER \cite{verma2023rnas}&WACV 23&		89.85	&90.78&-&	74.43	&76.20	&-&78.80&	82.35&-&	83.02&	85.11&\color{blue}{\textbf{90.29}}\\  
    
    $\mu$-BERT \cite{nguyen2023micron}&CVPR 23&90.34&	89.14	&-&\color{red}{\textbf{85.50}}&	\color{blue}{\textbf{83.84}}&-&	83.86&	84.75	&-&\color{blue}{\textbf{89.03}}	&88.42&-\\
   
    MFDAN \cite{cai2024mfdan}&TCSVT 24	&91.34&	93.26	&-&68.15&	70.43&-&	78.71&	81.96&-&	84.53	&86.88&-\\
    HTNet \cite{wang2024htnet}&NC 24	&	95.32	&95.16&-&	80.49&79.05&-&	81.31&	81.24	&-&86.03	&84.75&-\\
    SRMCL \cite{bao2024boosting}&TAC 24	&	96.35&96.49&-&	79.46	&80.53&-&	84.70&\color{blue}{\textbf{88.66}}	&-&86.30	&88.30&-\\
    TGMAE$^{\dagger}$ \cite{10687556}&ICME 24&	95.97&95.62&-&	82.25	&83.44&-&	84.15&78.85	&-&87.90	&88.02&-\\
    EDMDBN \cite{ma2025entire}&PRL 25 &94.84&96.19&-&	79.48	&80.85&-&	83.36&86.61&-&88.21	&\color{blue}{\textbf{89.33}}&-\\
HFA-Net\cite{zhang2025hfa}&CIS 25 &\color{blue}{\textbf{97.38}}&\color{blue}{\textbf{97.54}}&-&	77.86	&77.86&-&	\color{red}{\textbf{90.02}}&\color{red}{\textbf{89.38}}&-&88.00	&86.60&-\\
   \rowcolor{gray!20}

CausalNet&Proposed	&	\color{red}{\textbf{97.48}} \textsubscript{0.90}	&\color{red}{\textbf{97.82}} \textsubscript{1.05}&\color{red}{\textbf{98.21}\textsubscript{0.62}}
&	\color{blue}{\textbf{84.05}}\textsubscript{1.31}&\color{red}{\textbf{84.33}}\textsubscript{1.25}&\color{red}{\textbf{84.27}\textsubscript{1.32}}&	\color{blue}{\textbf{87.08}}\textsubscript{1.03}&85.22\textsubscript{1.15}&\color{red}{\textbf{90.83}\textsubscript{0.98}}&\color{red}{\textbf{89.30}}\textsubscript{0.72}	&\color{red}{\textbf{89.41}}\textsubscript{0.84}&\color{red}{\textbf{90.77}}\textsubscript{0.65}\\

 \bottomrule

\end{tabular}}
\end{table*}

\begin{table}[t]\centering

  \caption{Method Comparison on the MMEW for three-class task. - indicates that the results are not reported in corresponding papers.}
  \label{t2}
  \setstretch{1}
\resizebox{0.9\linewidth}{!}{
  \begin{tabular}{ccccc}
    \toprule
    \multirow{2}{*}{\textbf{Method}}&\multirow{2}{*}{\textbf{Pub-Yr}}&\multicolumn{3}{c}{\textbf{MMEW}}\\
    &&\textbf{UF1}&\textbf{UAR}&\textbf{ACC}\\
   
 \toprule
 LD-FMERN \cite{ni2023diverse}&KBS 23 &87.87&\color{blue}{\textbf{87.76}}&88.23\\
 EDMDBN \cite{ma2025entire}&PRL 25 &\color{blue}{\textbf{92.16}}&-&\color{blue}{\textbf{92.70}}\\
     \rowcolor{gray!20}

CausalNet&Proposed	&	\color{red}{\textbf{93.10}}\textsubscript{0.57}	&\color{red}{\textbf{92.15}}\textsubscript{0.48}&\color{red}{\textbf{94.44}}\textsubscript{0.30}\\ 
 \bottomrule

\end{tabular}}
\end{table}

\subsection{Comparision with SOTA Methods}
\label{s4.3}

The complete ME range input makes CausalNet more robust to key-frame errors. When the key-frame index is relatively accurate, it also shown competitive performance through CausalNet's learning of the causal relationship of ME in the onset-apex and apex-offset phases. With the annotated key frame index provided by ME benchmarks, we compare our method with the existing SOTA methods on three-class CDE task in \cref{t1} and three-class single dataset evaluation task on MMEW in \cref{t2} (see Supplementary Material for five-class results on MMEW). Note that all datasets, we analyze the performance by averaging the results over five runs. The results demonstrate that CausalNet has achieved a leading position in all metrics on the CASME II, composite, and MMEW datasets, and in some metrics on the SMIC dataset.

\subsection{Ablation Studies}

\begin{table}[t]\centering
\caption{Results of ablation studies of causal motion position learning module (CMPLM), spatial-temporal causal attention (STCA), causal relation minining (CRM) and pseudo-attention (PA). 
}
  \label{t3}
  \resizebox{1\linewidth}{!}{
  \begin{tabular}{cccccc}
  \toprule
 \multirow{2}{*}{\makecell{\textbf{CMPLM}}}&\multirow{2}{*}{\makecell{\textbf{STCA}}}& \multirow{2}{*}{\makecell{\textbf{CRM}}}&\multirow{2}{*}{\makecell{\textbf{PA}}}&\multicolumn{2}{c}{\textbf{Composite}}\\
&&&&\textbf{UF1}&\textbf{UAR}\\

\toprule
\color{black}{\usym{2717}}&\color{black}{\usym{2717}}&\color{black}{\usym{2717}}&\color{black}{\usym{2717}}&86.40\textsubscript{0.78}&86.55\textsubscript{0.69}\\
\color{black}{\usym{1F5F8}}&\color{black}{\usym{2717}}&\color{black}{\usym{2717}}&\color{black}{\usym{2717}}&87.01\textsubscript{0.82}&87.25\textsubscript{0.63}\\
\color{black}{\usym{1F5F8}}&\color{black}{\usym{1F5F8}}&\color{black}{\usym{2717}}&\color{black}{\usym{2717}}&88.42\textsubscript{0.65}&88.50\textsubscript{0.76}\\

\color{black}{\usym{1F5F8}}&\color{black}{\usym{1F5F8}}&\color{black}{\usym{1F5F8}}&\color{black}{\usym{2717}}&88.81\textsubscript{0.66}&88.94\textsubscript{0.52}\\
 \rowcolor{gray!20}
\color{black}{\usym{1F5F8}}&\color{black}{\usym{1F5F8}}&\color{black}{\usym{1F5F8}}&\color{black}{\usym{1F5F8}}&89.30\textsubscript{0.72}&89.41\textsubscript{0.84}\\
  \bottomrule
 \end{tabular}}
\end{table}

To analyze the functions of each component within CausalNet, we conduct ablation studies with the annotated key frames provided by the datasets. All the results are calculated from 5 independent runs, which are shown in \cref{t3}.

 \noindent\textbf{Causal Motion Position Learning Module}. After introducing it, the model improves UF1 by 0.61\% and UAR by 0.70\%, indicating that the module provides muscle movement position information to the main branch, reducing redundant attention and enhancing MER performance.

\noindent\textbf{Spatial-temporal Causal Attention}. After introducing causal and sparse attention, the model achieves an improvement of 1.41\% in UF1 and 1.25\% in UAR. The performance gain comes from local and global temporal-scale feature representations. Separating global and local information enables the model to focus on global ME motion while staying highly sensitive to local ME information.

 \noindent\textbf{Causal Relation Mining} boosts the model's performance by 0.39\% in UF1 and 0.44\% in UAR, verifying that the interaction between the two long-range features further enhances the relational learning of the entire ME.

 \noindent\textbf{Pseudo-attention} boosts the model's performance by 0.49\% in UF1 and 0.47\% in UAR, verifying that encoding absolute information contributes to the improvement of MER performance.

\section{Conclusions}
In this paper, we propose a novel framework using onset-apex OF and apex-offset OF as inputs to represent complete ME muscle movement for robust MER even with inaccurate key frame indexes. The causal motion position learning module and the causal attention block are proposed to enhance the attention of the network on key information to handle the whole ME sequence input range. Our method achieves robust MER under different levels of key frame index noise. Moreover, when using provided key-frames from the dataset, the proposed framework shows competitive performance in various standard ME benchmarks.

 \noindent\textbf{Limitations}. Although CausalNet demonstrates better robustness against key-frame errors, it still remains ineffective in handling cases where the expression lies entirely outside the onset-offset interval (i.e., $frame^{errors}_{offset}<frame^{GT}_{onset}$ or $frame^{GT}_{offset}<frame^{error}_{onset}$). In this situation, when large key-frame errors are present, further increasing the input range for MER provides limited improvement in recognition performance. Instead, optimizing the spotting algorithm or enhancing the training quality of manual annotations is likely to have a greater impact on improving current MER performance than refining the MER algorithm itself. These tasks will be explored in future work.

   \noindent\textbf{Acknowledgment}. 
This work was supported by the National Key R\&D Program of China under Grant 2022YFF0904305.

{

    \small
    \bibliographystyle{ieeenat_fullname}
    \bibliography{main}
}

\end{document}